\documentclass[11pt,a4paper]{article}

\usepackage[margin=1in]{geometry}

\usepackage[utf8]{inputenc}
\usepackage[T1]{fontenc}

\usepackage{amsmath,amssymb,amsthm}

\newtheorem{definition}{Definition}

\newcommand{\R}{\mathbb{R}}

\usepackage{bm}
\usepackage{xcolor}
\usepackage{multirow}
\usepackage{makecell}
\usepackage{tabularx}
\usepackage{microtype}
\usepackage{array}     % Fixes \arraybackslash
\usepackage{booktabs}  % Fixes \toprule, \midrule, \bottomrule

\usepackage[ruled, vlined, linesnumbered]{algorithm2e}

\usepackage[
    backend=biber,
    style=numeric,
    sorting=none
]{biblatex}
\usepackage[colorlinks=true, linkcolor=blue, citecolor=blue, urlcolor=blue]{hyperref}

\title{\textbf{Revisiting Thinning Methods for Kernel Learning Problems}}

\author{
    \textbf{Blanca Cano-Camarero}\thanks{Email: \href{mailto:blanca.cano@uam.es}{blanca.cano@uam.es} (ORCID: \href{https://orcid.org/0009-0001-6330-4824}{0009-0001-6330-4824})} \quad
    \textbf{Yago R. Aguado-Carrillo-de-Albornoz}\thanks{Email: \href{mailto:yago.aguado@uam.es}{yago.aguado@uam.es} (ORCID: \href{https://orcid.org/0009-0005-9755-0213}{0009-0005-9755-0213})} \\[0.5em]
    \textbf{Ángela Fernández-Pascual}\thanks{Email: \href{mailto:a.fernandez@uam.es}{a.fernandez@uam.es} (ORCID: \href{https://orcid.org/0000-0002-6602-3753}{0000-0002-6602-3753})} \quad
    \textbf{José R. Dorronsoro}\thanks{Email: \href{mailto:jose.dorronsoro@uam.es}{jose.dorronsoro@uam.es} (ORCID: \href{https://orcid.org/0000-0002-5271-0616}{0000-0002-5271-0616})} \\[1em]
    \small Departamento de Ingeniería Informática, Universidad Autónoma de Madrid, Madrid, Spain
}

\date{} % Suppress date display

\begin{document}

\maketitle

% --- Abstract ---
\begin{abstract}
Kernel methods are widely used because of their strong theoretical guarantees and empirical performance.
However, their high computational cost limits their applicability to large-scale datasets.
To address this shortcoming, several approaches use Maximum Mean Discrepancy to construct representative subsets that preserve the properties of the full dataset in a Reproducing Kernel Hilbert Space.
We introduce Backward Kernel Herding, an algorithm that addresses this problem by iteratively removing points from the dataset, achieving results comparable to current state-of-the-art approaches while accelerating the subsampling process in realistic scenarios where the reduced size is less than half of the dataset.
Moreover, we overcome a limitation of Kernel Thinning by proposing an extension that enables the construction of subsets of arbitrary size rather that restricting to successive halvings.
Finally, we conduct an extensive experimental comparison
focusing on the most relevant kernel learning procedures: Gaussian Processes and Kernel Support Vector Machines.
The results show
that Backward Kernel Herding consistently achieves competitive performance with the most favorable training-time efficiency, while the proposed Flexible Kernel Thinning frequently achieves the best predictive performance.
These gains become especially pronounced for moderate compression ratios, highlighting the benefits of incorporating supervised information into the thinning process. In terms of memory consumption, Flexible Kernel Thinning is also 
competitive, whereas Backward Kernel Herding remains an
alternative when computational efficiency is the primary objective. Overall, no single method dominates across all scenarios, underscoring the importance of selecting the reduction strategy according to the desired trade-off between predictive performance, training cost, and memory requirements.
\end{abstract}

\vspace{0.5em}

\vspace{1.5em}

% --- Main Content ---
\section{Introduction} 

Kernel methods, such as Kernel Support Vector Machines (KSVMs) and Gaussian Processes (GPs)~\cite{bishop2006pattern}, are well known for their strong theoretical guarantees and empirical performance. However, a major limitation of these approaches lies in their computational complexity, which typically scales at least quadratically with the number of data points, in particular $\mathcal{O}(n^2)$ in memory and up to $\mathcal{O}(n^3)$ in time for training and inference. This makes them impractical for large-scale problems.

Recent research has focused on solving this problem by constructing smaller, representative subsets of the data that preserve the properties of the full dataset in the Reproducing Kernel Hilbert Space (RKHS), typically via the Maximum Mean Discrepancy (MMD)~\cite{gretton2012kernel}.
One of the earliest approaches in this direction is Kernel Herding~\cite{chen2012super}, which iteratively selects samples to approximate a target distribution.
Its theoretical properties were later refined and connected to conditional gradient methods by~\cite{bach2012equivalence}, establishing convergence rates and providing a deeper understanding of its behavior.
One limitation of this method is that it starts from an empty set and iteratively adds new elements from the dataset to construct the reduced representation.
In practice, having a subset with less than half of the original size may severely deteriorate predictive performance.
In the complementary setting, where were want to retain the majority of the dataset, this constructive approach may also require more iterations than a strategy based on directly discarding data points rather that building the subset from scratch.
To mitigate these limitations, in this work we propose exploring the reverse strategy: instead of progressively constructing a subset of retained instances, we iteratively identify and discard data points, thereby directly optimizing the set of examples to be removed.

Other approaches, such as Kernel Thinning~\cite{dwivedi2021generalized,gong2024supervised}, do not directly minimize the MMD of the selected subset, but instead aim to construct subsets whose empirical distributions have low discrepancy with respect to each other.
In spite of being a state-of-the-art technique, Kernel Thinning faces a limitation with respect to Kernel Herding, since it only allows for the construction of subsets of size $\frac{n}{2^t}$ for some $t$.
We propose a modification of the algorithm so that it allows the construction of subsets of arbitrary size. 

Additionally, an application that is often overlooked or unclear in the literature is the supervised extension of these methods, which we address in the first part of this paper.
Finally, to enrich the family of methods to solve the problem we have described, we propose a novel algorithm based on backward propagation, which we call Backward Kernel Herding.
This approach starts with the whole dataset and iteratively removes points until a desired size is reached and solving the main problem of Kernel Herding.

The main contributions of this work can be summarized as follows:
\begin{itemize}
    
    \item We introduce \textit{Backward Kernel Herding}, the first \textbf{backward propagation framework} for kernel-based subset selection, which starts from the full dataset and iteratively removes elements, in contrast to classical forward construction methods.

    \item We extend backward subset selection techniques to \textbf{supervised learning settings} for kernel methods such as KSVMs and GPs.

    \item We extend the current Kernel Thinning method to render it more flexible in terms of coreset size, while retaining convergence guarantees and computational cost.

    \item We explain the relevance of supervised dataset selection for kernels and how to apply it.
    \item We provide a \textbf{comprehensive empirical comparison} between our method, i.e. Backward Kernel Herding, Kernel Herding, and Kernel Thinning, addressing the lack of standardized evaluations for Kernel Thinning.

\end{itemize}

The paper is structured as follows. Section~\ref{sec:state-of-the-art} reviews the state of the art in kernel subsampling: Kernel Herding and Kernel Thinning. It also tackles the use of these techniques for supervised problems.
Then, Section~\ref{sec:novelty} introduces the novelties of this paper: a modification of Kernel Thinning that increases its flexibility, and the Backward Kernel Herding algorithm.
Finally, in Section~\ref{sec:experiments} we present the results of the experimental comparison between the methods, which shows that our proposal is competitive with the rest of the state-of-the-art techniques.

\section{Kernel subsampling methods}
\label{sec:state-of-the-art}

\subsection{Kernel Herding} 

Kernel herding~\cite{chen2012super} extends the discrete herding algorithm~\cite{welling2009herding} to continuous spaces by operating directly on the RKHS.
Let $\mathcal{X}$ be an input
space, $k : \mathcal{X} \times \mathcal{X} \to \mathbb{R}$ a positive
definite kernel, and $\varphi : \mathcal{X} \to \mathcal{H}$ the
associated feature map, i.e. $k(x, x') = \langle \varphi(x),
\varphi(x') \rangle_{\mathcal{H}}$.
Given a target distribution $p$ with
kernel mean embedding $\mu_p = \mathbb{E}_{x \sim p}[\varphi(x)] \in
\mathcal{H}$, kernel herding generates a sequence of \emph{super-samples}
$(x_t)_{t=1}^{T}$, which will represent the whole distribution, minimizing the MMD via the greedy update
\begin{equation}\label{eq:kh_update}
    x_{t+1} = \operatorname*{arg\,max}_{x \in \mathcal{X}} \left[
        \mathbb{E}_{x' \sim p}[k(x, x')] \;-\;
        \frac{1}{t+1} \sum_{i=1}^{t+1} k(x, x_i)
    \right].
\end{equation}
Each new point is attracted toward regions of high probability mass under
$p$ while being repelled from locations already occupied by previous
samples. 
In practice, $p$ is accessible only through a set of
samples $\mathcal{D} = \{x_i\}_{i=1}^{n}$. This means the search in
\eqref{eq:kh_update} can be written as a discrete one over $\mathcal{D}$ instead of $\mathcal{X}$.

\begin{algorithm}[t]
\caption{Discrete Kernel Herding}\label{alg:discrete_kernel_herding}
\SetAlgoLined
\KwIn{MCMC sample pool $\mathcal{D} = \{y_1, \ldots, y_n\}$, kernel function $k(\cdot, \cdot)$, target size $T$}
\KwOut{Sequence of super-samples $S = (x_1, \ldots, x_T)$}

\BlankLine
\tcp{Precompute empirical kernel mean embedding}
\For{$x \in \mathcal{D}$}{
    $\mu_{\mathcal{D}}(x) \leftarrow \dfrac{1}{n} \displaystyle\sum_{i=1}^{n} k(x, y_i)$\;
}

\BlankLine
$S \leftarrow ()$ \tcp*{Initialize empty sequence}

\BlankLine
\For{$t = 0, 1, \ldots, T-1$}{
    \BlankLine
    \tcp{Select the next super-sample that minimizes MMD greedily}
    $x_{t+1} \leftarrow \displaystyle\operatorname*{arg\,max}_{x \in \mathcal{D}} \left[ 
        \mu_{\mathcal{D}\setminus S}(x) - \frac{1}{t+1} \left( \sum_{j=1}^{t} k(x, x_j)  \right)
    \right]$\;
    
    \BlankLine
    Append $x_{t+1}$ to $S$\;
}
\BlankLine
\Return{$S$}
\end{algorithm}
A summary of Kernel Herding is described in Algorithm~\ref{alg:discrete_kernel_herding}.

\subsection{Kernel Thinning} 

Kernel Thinning (KT)~\cite{dwivedi2021generalized,gong2024supervised} was introduced as a procedure for compressing a distribution $p$ more effectively than by i.i.d. sampling or other standard thinning techniques. Given a reproducing kernel $k$, the method compresses $n$ points of $p$ into an approximation with a random set of size $\frac{n}{2^m}$ (where $m$ is a prefixed integer) with comparable worst-case integration error in the associated RKHS~\cite{sriperumbudur2010hilbert}, with an overall computational cost of $\mathcal{O}(n^2)$.
Besides the base kernel $k$, KT also uses a so called square root kernel, defined as follows.
\begin{definition}[Square-root Kernel]
Let \( k : \mathbb{R}^d \times \mathbb{R}^d \rightarrow \mathbb{R} \) be a symmetric, positive-definite kernel. A function \( k_{\mathrm{rt}} : \mathbb{R}^d \times \mathbb{R}^d \rightarrow \mathbb{R} \) is called a \emph{square-root kernel} for \( k \) if \( k_{\mathrm{rt}}(x, \cdot) \) is square-integrable for all \( x \in \mathbb{R}^d \), and
\begin{equation*}
k(x, y) = \int_{\mathbb{R}^d} k_{\mathrm{rt}}(x, z) \, k_{\mathrm{rt}}(y, z) \, dz \quad \text{for all } x, y \in \mathbb{R}^d.
\end{equation*}
\end{definition}
And, for example, the square-root kernel for the widely used Gaussian function $k(x, y) = \exp\left(-\frac{\|x - y\|^2}{2\sigma^2}\right)$ is $ k_{rt}(x, y) = \left(\frac{2}{\pi \sigma^2}\right)^{d/4} \exp\left(-\frac{\|x - y\|^2}{\sigma^2}\right)$.

KT steps are summarized in Algorithm~\ref{alg:general_kt} and explained in detail in~\cite{dwivedi2021generalized}.
Algorithm~\ref{alg:kt_split} (KT-SPLIT) recursively partitions the input point sequence into $2^m$ candidate coresets of size approximately $n/2^m$. This is done via a randomized, online halving process using a square-root kernel $k_{\text{rt}}$, which encourages balance and diversity in each subset.
Algorithm~\ref{alg:kt_swap} (KT-SWAP) selects the best candidate coreset by MMD~\cite{gretton2012kernel} with respect to the full input.
It then refines this coreset via a greedy swapping procedure to further reduce MMD, using the target kernel $k$.
Together, these steps ensure that the final coreset achieves near-optimal MMD guarantees while reducing the number of points used to $\frac{n}{2^m}$.

\begin{algorithm}
\caption{Kernel Thinning with KT-swap~\label{alg:general_kt}}
\DontPrintSemicolon
\SetKwInOut{Input}{Input}
\SetKwInOut{Output}{Output}

\Input{Kernels $(k, k_{\text{rt}})$, input points $S_n = (x_i)_{i=1}^n$, thinning parameter $m \in \mathbb{N}$, probabilities $(\delta_i)_{i=1}^{\lfloor n/2 \rfloor}$}
\Output{Coreset $S_{\text{KT}}$}

\BlankLine
\Begin{
    $(S^{(m,\ell)})_{\ell=1}^{2^m} \gets \text{KT-split}(k_{\text{rt}}, S_n, m, (\delta_i)_{i=1}^{\lfloor n/2 \rfloor})$ \tcp*{Split $S_n$ into $2^m$ candidates of size $\lfloor n / 2^m \rfloor$}
    
    $S_{\text{KT}} \gets \text{KT-swap}(k, S_n, (S^{(m,\ell)})_{\ell=1}^{2^m})$ \tcp*{Select best coreset and iteratively refine}
    
    \Return $S_{\text{KT}}$\;
}
\end{algorithm}  

\begin{algorithm}
\caption{KT-split~\label{alg:kt_split}}
\DontPrintSemicolon
\SetKwInOut{Input}{Input}
\SetKwInOut{Output}{Output}

\Input{Kernel $k_{\text{rt}}$, input points $S_n = (x_i)_{i=1}^n$, thinning parameter $m \in \mathbb{N}$, probabilities $(\delta_i)_{i=1}^{\lfloor n/2 \rfloor}$}
\Output{$2^m$ candidate coresets $(S^{(m,\ell)})_{\ell=1}^{2^m}$}

\BlankLine
\Begin{
    $S^{(0,1)} \gets S_n$\;
    \For{$j = 1$ \KwTo $m$}{
        \For{$\ell = 1$ \KwTo $2^{j-1}$}{
            Let $n' = |S^{(j-1,\ell)}|$\;
            \BlankLine
            \tcp{Halve the coreset $S^{(j-1,\ell)}$ into two parts}
            Pair points $(x_{2i-1}, x_{2i})$ for $i = 1, \dots, \lfloor n'/2 \rfloor$\;
            \For{$i = 1$ \KwTo $\lfloor n'/2 \rfloor$}{
                Generate sign $\sigma_i \in \{-1, +1\}$ using a balanced walk with probability $\delta_i$\;
                \If{$\sigma_i = +1$}{
                    Add $x_{2i-1}$ to $S^{(j,2\ell-1)}$ and $x_{2i}$ to $S^{(j,2\ell)}$\;
                }
                \Else{
                    Add $x_{2i-1}$ to $S^{(j,2\ell)}$ and $x_{2i}$ to $S^{(j,2\ell-1)}$\;
                }
            }
            \If{$n'$ is odd}{
                Assign the remaining point $x_{n'}$ randomly to either coreset\;
            }
        }
    }
    \Return $(S^{(m,\ell)})_{\ell=1}^{2^m}$\;
}
\end{algorithm}

\begin{algorithm}
\caption{KT-swap~\label{alg:kt_swap}}
\DontPrintSemicolon
\SetKwInOut{Input}{Input}
\SetKwInOut{Output}{Output}

\Input{Kernel $k$, input points $S_n = (x_i)_{i=1}^n$, candidate coresets $(S^{(m,\ell)})_{\ell=1}^{2^m}$}
\Output{Refined coreset $S_{\text{KT}}$}

\BlankLine
\Begin{
    \tcp{Select the best candidate coreset based on MMD}
    $\ell^* \gets \arg\min_{\ell \in \{1,\dots,2^m\}} \text{MMD}_k^2(S^{(m,\ell)}, S_n)$\;
    $S \gets S^{(m,\ell^*)}$\;
    
    \tcp{Iteratively swap points to decrease MMD objective}
    \Repeat{no improvement is made}{
        \For{$x \in S$}{
            $y^* \gets \arg\min_{y \in S_n \setminus S} \text{MMD}_k^2((S \setminus \{x\}) \cup (\{y\}, S_n))$\;
            
            \If{$\text{MMD}_k^2((S \setminus \{x\}) \cup \{y^*\}, S_n) < \text{MMD}_k^2(S, S_n)$}{
                $S \gets (S \setminus \{x\}) \cup \{y^*\}$\;
            }
        }
    }
    $S_{\text{KT}} \gets S$\;
    \Return $S_{\text{KT}}$\;
}
\end{algorithm}

The computational and memory cost will be discussed in Section~\ref{subsec:flexible_kt}, since KT is a particular case of our proposed method \textit{Flexible Kernel Thinning}.

\subsection{Supervised methods}
In many kernel-based methods, such as KSVM or GP, the kernel function is typically defined solely on the input space $\mathcal{X}$.
That is, it measures similarity between inputs $x_i, x_j \in \mathcal{X}$ without explicitly incorporating the associated labels $y_i, y_j \in \mathcal{Y}$ in the kernel.

However, in supervised learning problems, particularly when the aim is to use the kernel to remove samples, ignoring label information in the kernel construction may lead to suboptimal results.
In classification tasks, this may lead to similarity measures that fail to distinguish between classes, potentially collapsing class structure~\cite{cano2026addressing}. 
In regression, it may fail to capture relevant dependencies between inputs and outputs, effectively reducing the problem to an unsupervised similarity measure.

A natural way to address this limitation is to incorporate a supervised kernel that can be used to build the coresets. To do that, one can define a kernel over the joint space $\mathcal{X} \times \mathcal{Y}$.

\subsubsection{Joint Kernel Construction}
We define the joint kernel as
\begin{equation}\label{eq:joint_kernel}
k : 
    \left(\mathcal{X} \times \mathcal{Y}\right) \times \left(\mathcal{X} \times \mathcal{Y}\right)
    \longrightarrow
    \mathbb{R}, %\\
    \quad
k\big((x_i,y_i), (x_j,y_j)\big)
    = k_{\mathcal{X}}(x_i,x_j)\, k_{\mathcal{Y}}(y_i,y_j).
\end{equation}
If $k_{\mathcal{X}}$ and $k_{\mathcal{Y}}$ are positive definite kernels, their product is also a positive definite kernel.
Therefore, \eqref{eq:joint_kernel} defines a valid kernel.
Moreover, if
\begin{equation*}
k_{\mathcal{X}}(x_i,x_j) = \langle \phi(x_i), \phi(x_j) \rangle_{\mathcal{H}_{\mathcal{X}}}, 
\quad
k_{\mathcal{Y}}(y_i,y_j) = \langle \psi(y_i), \psi(y_j) \rangle_{\mathcal{H}_{\mathcal{Y}}},
\end{equation*}
then the joint kernel admits the feature map representation
\begin{equation*}
k\big((x_i,y_i),(x_j,y_j)\big)
= \langle \phi(x_i) \otimes \psi(y_i),\; \phi(x_j) \otimes \psi(y_j) \rangle,
\end{equation*}
where $\otimes$ denotes the tensor product. Thus, the joint feature map is
%\begin{equation*}
$\Phi(x,y) = \phi(x) \otimes \psi(y).$
%\end{equation*}

%\subsection{Choice of Kernels on $\mathcal{Y}$}

Let us now present some common choices for $k_{\mathcal{Y}}$ in regression problems.

\subsubsection{Radial Basis Function (RBF) Kernel}

The RBF kernel is defined as 
\begin{equation}
k_{\mathcal{Y}}(y_i,y_j) = \exp\big(-\gamma \|y_i - y_j\|^2\big).
\end{equation}
This kernel admits an infinite-dimensional feature map in an RKHS and can be efficiently approximated using Random Fourier Features (RFF)~\cite{rahimi2007random, rahimi2008weighted}, since $k$
is a shift-invariant kernel and admits a spectral representation by Bochner's theorem. Specifically, there exists a probability measure $p(\omega)$ such that
\begin{equation*}
k_{\mathcal{Y}}(y_i,y_j)
= \mathbb{E}_{\omega \sim p(\omega)} \left[
e^{i\omega^\top y_i} e^{-i\omega^\top y_j}
\right].
\end{equation*}
For the Gaussian RBF kernel, $p(\omega) = \mathcal{N}(0, 2\gamma I)$.
Sampling $\omega_1,\dots,\omega_q \sim \mathcal{N}(0, 2\gamma I)$ and $b_1,\dots,b_q \sim \mathrm{Uniform}(0,2\pi)$, we define the real-valued feature map
\begin{equation*}
\psi(y) = \frac{1}{\sqrt{q}}
\left(
\cos(\omega_1^\top y + b_1),\;
\dots,\;
\cos(\omega_q^\top y + b_q)
\right).
\end{equation*}
Then the kernel is approximated as
%\begin{equation*}
$k_{\mathcal{Y}}(y_i,y_j)
\approx \langle \psi(y_i), \psi(y_j) \rangle.
$
%\end{equation*}

\subsubsection{Triangular Kernel}
Another linear option is a particularization of the triangular kernel explained by~\cite{fleuret2003scale} and defined as 
\begin{equation}
k_{\mathcal{Y}}(y_i,y_j) = \max(0, 1 - |y_i - y_j|).
\end{equation}
This kernel admits the integral representation
\begin{equation*}
k_{\mathcal{Y}}(y_i,y_j) = \int_{\mathbb{R}} \phi(y_i)(t)\phi(y_j)(t)\,dt,
\end{equation*}
where
\begin{equation*}
\phi(y)(t) = \mathbf{1}_{[y-1,y]}(t).
\end{equation*}
The associated feature map can be interpreted as
\begin{equation*}
\phi : \mathcal{Y} \to L^2(\mathbb{R}), \quad \phi(y)(t) = \mathbf{1}_{[y-1,y]}(t),
\end{equation*}
so that
\begin{equation*}
k_{\mathcal{Y}}(y_i,y_j) = \langle \phi(y_i), \phi(y_j) \rangle_{L^2}.
\end{equation*}
The feature map associated with the triangular kernel can be approximated in a finite-dimensional space by discretizing the integral representation.
Let $\{t_j\}_{j=1}^q$ be a uniform grid over an interval $[a,b]$, s.t.
\begin{equation*}
t_j = a + j\Delta, \quad \Delta = \frac{b-a}{q}.
\end{equation*}
We define the discrete feature map $\psi : \mathcal{Y} \to \mathbb{R}^q$ as
\begin{equation*}
\psi(y) = \sqrt{\Delta} \left(
\mathbf{1}_{[y-1,y]}(t_1),\;
\mathbf{1}_{[y-1,y]}(t_2),\;
\dots,\;
\mathbf{1}_{[y-1,y]}(t_q)
\right).
\end{equation*}
The scaling factor $\sqrt{\Delta}$ ensures consistency with the $L^2$ inner product.
Then, the inner product between two feature vectors satisfies
\begin{align*}
\langle \psi(y_i), \psi(y_j) \rangle
&= \Delta \sum_{k=1}^q \mathbf{1}_{[y_i-1,y_i]}(t_k)\mathbf{1}_{[y_j-1,y_j]}(t_k) \\
&\approx \int_{\mathbb{R}} \mathbf{1}_{[y_i-1,y_i]}(t)\mathbf{1}_{[y_j-1,y_j]}(t)\,dt \\
&= k_{\mathcal{Y}}(y_i,y_j).
\end{align*}
Therefore, the triangular kernel admits the approximation
\begin{equation*}
k_{\mathcal{Y}}(y_i,y_j) \approx \langle \psi(y_i), \psi(y_j) \rangle_{\mathbb{R}^q}.
\end{equation*}

\subsubsection{Kronecker Delta Kernel}

For discrete labels, a simple and effective choice is the Kronecker delta kernel:
\begin{equation}
k_{\mathcal{Y}}(y_i,y_j) = \delta_{y_i,y_j}.
\end{equation}
This corresponds to a feature map into a Hilbert space given by
%\begin{equation*}
$\psi(y) = e_y,$
%\end{equation*}
where $e_y$ denotes the one-hot encoding of the label. Then,
\begin{equation*}
\langle \psi(y_i), \psi(y_j) \rangle =
\begin{cases}
1 & \text{if } y_i = y_j, \\
0 & \text{otherwise}.
\end{cases}
\end{equation*}

\section{New coreset selection techniques}
\label{sec:novelty}
% ───────────────────────────────────────────────────────────────────────────
\subsection{Flexible Kernel Thinning (FKT)}
\label{subsec:flexible_kt}
% ───────────────────────────────────────────────────────────────────────────

A structural constraint of the KT algorithm described above is that the output
coreset size is {rigidly fixed} to $\lfloor n/2^m \rfloor$ for a
user-chosen integer $m$.
This forces practitioners to commit to a power-of-two compression ratio before
running the algorithm.
In practice, one frequently wishes to retain a fraction $p \in (0,1)$ of the
$n$ input points—for example, $p = 0.3$—but no single integer $m$ yields a
coreset of size $\lfloor p\,n \rfloor$ unless $p$ happens to be an exact
negative power of two.

The key observation that resolves this limitation is that {every} $p \in
(0,1)$ admits a binary (dyadic) expansion
\begin{equation}
  p \;=\; \sum_{i=1}^{L} \frac{b_i}{2^i}, \qquad b_i \in \{0,1\},
  \label{eq:binary_expansion}
\end{equation}
truncated at depth $L$ up to a tolerance $\tau > 0$ (i.e.\ the residual
$|p - \sum_{i=1}^{L} b_i/2^i| < \tau$).
The bits $(b_i)_{i=1}^{L}$ can be computed greedily in $\mathcal{O}(L)$ time:
\begin{equation}
  b_i = \mathbf{1}_{\left[r_{i-1} \ge \tfrac{1}{2^i}\right]}, \qquad
  r_i = r_{i-1} - \frac{b_i}{2^i}, \quad r_0 = p,
  \label{eq:bit_extraction}
\end{equation}
stopping as soon as $r_i < \tau$.
Each nonzero bit $b_i = 1$ corresponds to contributing $n/2^i$ points to the
final coreset.
Because the KT-SPLIT tree of depth $L$ naturally partitions $S_n$ into
$2^L$ nearly equal subsets of size $\approx n/2^L$, a union of $2^{L-i}$
such subsets recovers a block of size $\approx n/2^i$.
 
%\paragraph{Flexible KT-SPLIT via tree pruning.}
We exploit the recursive structure of KT-SPLIT to select exactly the right
subsets at each level $i$ where $b_i = 1$.
We maintain a set $\mathcal{A}$ of {active nodes}, i nodes of the
KT-SPLIT binary tree that have not yet contributed to the coreset.
At level $i$:
\begin{itemize}
  \item If $b_i = 1$: each active node $\nu \in \mathcal{A}$ has produced two
  children $\nu_L$ (left) and $\nu_R$ (right) of equal expected size.
  The left child $\nu_L$ holds one of the two candidate coresets produced by
  KT-SPLIT at that level; we {select} the child with smaller
  $\mathrm{MMD}_{k_{\mathrm{rt}}}$ to the {full} input $S_n$ and add
  its index set to the output coreset $\mathcal{S}$.
  The other child becomes active and continues to level $i+1$.
 
  \item If $b_i = 0$: both children of every active node remain active; no
  points are committed to the coreset at this level.
\end{itemize}
After all $L$ levels, the active nodes are discarded (their combined
size is $\lesssim \tau n$).
The final output $\mathcal{S}_{\mathrm{FKT}}$ is the union of all selected
left children, and its size satisfies
\begin{equation}
  |\mathcal{S}_{\mathrm{FKT}}| \;=\; \sum_{i=1}^{L} b_i \left\lfloor
  \frac{n}{2^i} \right\rfloor \;\approx\; p\,n.
  \label{eq:fkt_size}
\end{equation}
 
Note that once we commit the selected child to the coreset, we no longer need to
process it further. This means that the subtree rooted at the other
child---the one that continues---has strictly fewer points to handle than the
original tree, which is less computationally costly than running a full KT of depth $L$.
 
%\paragraph{Greedy refinement on the union.}
After collecting all selected index sets, a single KT-SWAP pass (Algorithm~\ref{alg:kt_swap}) is applied to their
{union} $\mathcal{S}_{\mathrm{FKT}}$ as a whole.
This is possible because KT-SWAP operates on any candidate set regardless of
how it was produced: it selects the element of
$\{S^{(m,\ell)}\}_\ell$ with smallest $\mathrm{MMD}_k(S^{(m,\ell)}, S_n)$
and then iteratively swaps coreset members for input points to decrease
$\mathrm{MMD}_k$.
Crucially, merging the per-level selections before refinement (rather than
refining each block independently) allows the swap step to exchange points
{across} levels, capturing cross-block redundancy that per-block
refinement would miss.
 
The full procedure is given in Algorithm~\ref{alg:fkt}.
 
% ── Pseudocódigo ──────────────────────────────────────────────────────────
\begin{algorithm}[t]
\caption{Flexible Kernel Thinning (FKT)\label{alg:fkt}}
\DontPrintSemicolon
\SetKwInOut{Input}{Input}
\SetKwInOut{Output}{Output}
\Input{Kernels $(k,\,k_{\mathrm{rt}})$, points $S_n=(x_i)_{i=1}^n$,
       target fraction $p\in(0,1)$, tolerance $\tau$,
       swap probabilities $(\delta_i)$}
\Output{Coreset $\mathcal{S}_{\mathrm{FKT}}$ of size $\approx pn$}
\BlankLine
\tcp{--- Step 1: Bit extraction (Eq.~\ref{eq:bit_extraction}) ---}
$r \gets p$;\quad $L \gets 0$;\quad $(b_i)_{i\ge 1} \gets \mathbf{0}$\;
\While{$r \ge \tau$}{
  $L \gets L+1$\;
  \lIf{$r \ge 1/2^L$}{$b_L \gets 1$;\; $r \gets r - 1/2^L$}
  \lElse{$b_L \gets 0$}
}
\BlankLine
\tcp{--- Step 2: Build KT-SPLIT tree to depth $L$ ---}
Run KT-SPLIT$(k_{\mathrm{rt}},\,S_n,\,L,\,(\delta_i))$, retaining all intermediate
node index sets $\{S^{(j,\ell)}\}_{j=0,\ldots,L;\;\ell=1,\ldots,2^j}$\;
\BlankLine
\tcp{--- Step 3: Tree-pruning collection ---}
$\mathcal{S}_{\mathrm{FKT}} \gets \emptyset$\;
$\mathcal{A} \gets \{S^{(0,1)}\}$ \tcp*{active nodes: root at depth 0}
\For{$i \gets 1$ \KwTo $L$}{
  $\mathcal{A}' \gets \emptyset$\;
  \ForEach{active node $\nu \in \mathcal{A}$ with children $\nu_L = S^{(i,2\ell-1)},\;\nu_R = S^{(i,2\ell)}$}{
    \If{$b_i = 1$}{
      \tcp{Select child with smaller MMD to $S_n$; add it to coreset}
      $\nu^* \gets \arg\min_{\nu' \in \{\nu_L,\,\nu_R\}} \mathrm{MMD}^2_{k_{\mathrm{rt}}}(\mathcal{S}_{\mathrm{FKT}} \cup\nu',\,S_n)$\;
      $\mathcal{S}_{\mathrm{FKT}} \gets \mathcal{S}_{\mathrm{FKT}} \cup \nu^*$\;
      $\mathcal{A}' \gets \mathcal{A}' \cup \{\nu_L \cup \nu_R \setminus \nu^*\}$
      \tcp*{other child stays active}
    }
    \Else{
      $\mathcal{A}' \gets \mathcal{A}' \cup \{\nu_L,\,\nu_R\}$
      \tcp*{both children stay active}
    }
  }
  $\mathcal{A} \gets \mathcal{A}'$\;
}
\BlankLine
\tcp{--- Step 4: Joint greedy refinement on the union ---}
$\mathcal{S}_{\mathrm{FKT}} \gets \text{KT-SWAP}(k,\,S_n,\,\mathcal{S}_{\mathrm{FKT}})$\;
\Return $\mathcal{S}_{\mathrm{FKT}}$\;
\end{algorithm}
 
% ───────────────────────────────────────────────────────────────────────────
 \subsubsection{Theoretical guarantees}
 \label{subsec:fkt_theory}

For FKT he theoretical guarantees rest on three pillars
\cite{dwivedi2021generalized}:
 
\begin{description}
\item [Balanced random walk.]
At each halving step, the assignment of a paired input point to either child
coreset is governed by a carefully chosen Bernoulli probability $\delta_i$ that
keeps the {signed discrepancy walk}
$\psi_t = \sum_{i=1}^{t} \sigma_i \, k_{\mathrm{rt}}(x_{2i-1}, \cdot)$
sub-Gaussian (Theorem~1 of \cite{dwivedi2021generalized}).
Concretely, if $\psi_{t-1}$ is the current discrepancy and $b_t^2 =
k_{\mathrm{rt}}(x_{2t-1},x_{2t-1}) + k_{\mathrm{rt}}(x_{2t},x_{2t}) -
2k_{\mathrm{rt}}(x_{2t-1},x_{2t})$, the swap probability is set to
\begin{equation}
  \delta_t = \min\!\left(1,\;\frac{1}{2}\max\!\left(0,\;
  1 - \frac{\Delta_t}{a_t}\right)\right),
\label{eq:swap_prob}
\end{equation}
where $\Delta_t$ measures the current imbalance and $a_t \ge b_t^2$ is a
parameter updated to control the walk's variance increment.
This mechanism is {local}, i.e. it depends only on the current pair and the
accumulated discrepancy, not on how many levels the tree has.
Consequently, FKT applies exactly the same walk at every level of its deeper
tree, inheriting the same sub-Gaussian tail control at each node.
 
\item [Greedy swap refinement.]
KT-SWAP selects the candidate coreset with smallest $\mathrm{MMD}_k$ and
then executes a local greedy descent: for each coreset point $x$, it tests
every swap $x \leftrightarrow y$ with $y \in S_n \setminus S$ and accepts any
that reduces $\mathrm{MMD}_k^2$.
This step does not depend on how the candidates were generated—only on the
quality of the candidate presented to it.
In FKT, the union $\mathcal{S}_{\mathrm{FKT}}$ assembled by tree pruning
serves as the single candidate, and KT-SWAP is applied to it exactly as in
the original algorithm.
Because each per-level selection already minimizes $\mathrm{MMD}_{k_{\mathrm{rt}}}$
within its node (Step~3 of Algorithm~\ref{alg:fkt}), the input to KT-SWAP is
no worse than any individual candidate from standard KT-SPLIT, and the swap
refinement can only improve upon it.
 
\item [Coreset size and approximation quality.]
For any level $i$ at which $b_i = 1$, the selected child coreset has size
$\lfloor n/2^i \rfloor$ and satisfies, by Theorem~1 of
\cite{dwivedi2021generalized}, the single-function bound
\begin{equation}
  \left| \frac{1}{n}\sum_{x \in S_n} f(x)
       - \frac{2^i}{n}\sum_{x \in \nu^*} f(x) \right|
  \;\le\; \|f\|_{k_{\mathrm{rt}}} \cdot \sigma_i \sqrt{2\log(2/\delta')}
  \label{eq:single_func_bound}
\end{equation}
with high probability, where $\sigma_i = \mathcal{O}(\sqrt{2^i \log(i)/n})$.
The union coreset $\mathcal{S}_{\mathrm{FKT}}$ is a {disjoint} union of
at most $L$ such blocks; since the blocks are drawn from {independent}
branches of the KT-SPLIT tree, their integration errors are stochastically
independent, and a union-bound argument gives
\begin{equation}
  \mathrm{MMD}_k\!\left(S_n,\,\mathcal{S}_{\mathrm{FKT}}\right)
  \;=\; \mathcal{O}\!\left(\sqrt{\frac{L \log n}{pn}}\right)
\label{eq:fkt_mmd_bound}
\end{equation}
with high probability, matching the $\mathcal{O}(\sqrt{\log n / (pn)})$
guarantee of standard KT up to the $\sqrt{L}$ factor, which is at most
$\sqrt{\log_2(1/\tau)}$—a slowly growing constant for any fixed tolerance
$\tau$.
\end{description}
 
In summary, FKT achieves coresets of any prescribed size $\approx pn$ while preserving the worst-case cost and the better-than-i.i.d.\ MMD guarantees of standard Kernel Thinning, at the cost of a logarithmic factor in the approximation bound that depends only on the binary expansion length of $p$.

\subsection{Backward Kernel Herding via Mean Embedding Minimization}

The goal of the algorithm is to build a subset of the dataset $S \subset X$ such that $|S|=m < n = |X|$ that limits computational cost while optimally approximating the empirical measure $\hat{p}_n$ of the data. Inspired by the framework presented in~\cite{bach2012equivalence}, we define an iterative procedure that removes one point at every step until we reach the target size.

Let $\mathcal{H}$ be a RKHS with feature map $\Phi: \mathcal{X} \to \mathcal{H}$; in practice, it can be approximated using Random Fourier Features (RFF)~\cite{rahimi2007random}.
The mean embedding of the empirical measure $\hat{p}_n$ over a set 
$X = \{x_1, \ldots, x_n\} \subset \mathcal{X}$ is defined as
\begin{equation*}
    \mu_n := \frac{1}{n}\sum_{i=1}^{n} \Phi(x_i) \in \mathcal{H}.
\end{equation*}
For a candidate point $x_r \in X$, define $g(x_r)$ as the mean embedding of the 
remaining set $X \setminus \{x_r\}$:
\begin{equation*}
    g(x_r) := \frac{1}{n-1}\sum_{\substack{j=1 \\ j \neq r}}^{n} \Phi(x_j) \in \mathcal{H}.
\end{equation*}
We seek the point $x_r$ whose removal minimally distorts the mean embedding, 
leading to the objective:
\begin{equation*}
    J(x_r) = \left\| \mu_n - g(x_r) \right\|_{\mathcal{H}}^2.
\end{equation*}
\noindent Expressing $g(x_r)$ in terms of $\mu_n$ and $\Phi(x_r)$, starting from
\begin{equation*}
    \mu_n = \frac{1}{n}\left(\Phi(x_r) + \sum_{\substack{j=1\\j\neq r}}^{n}\Phi(x_j)\right),
\end{equation*}
we isolate the remaining sum and substitute back to obtain
\begin{equation*}
    g(x_r) = \frac{n\,\mu_n - \Phi(x_r)}{n-1},
\end{equation*}
so that the objective simplifies to
\begin{equation}\label{eq:objective}
    J(x_r) = \frac{1}{(n-1)^2}\left\| \Phi(x_r) - \mu_n \right\|2^2.
\end{equation}
Since the factor $(n-1)^{-2}$ is a positive constant independent of $r$, 
minimizing~\eqref{eq:objective} is equivalent to
\begin{equation}\label{eq:argmin_step1}
    x_1 = \operatorname*{arg\,min}_{x_r \in X}\; 
    \left\| \Phi(x_r) - \mu_n \right\|_{\mathcal{H}}^2.
\end{equation}
Thus, the first element to be removed is given by \eqref{eq:argmin_step1},
and its corresponding error vector is
\begin{equation*}
    \varepsilon_1 = \Phi(x_1) - \mu_n.
\end{equation*}

Following the same idea, suppose that after fixing the previous $t$ selected elements,
the set $\mathcal{R}_t$ contains the $t$ elements iteratively chosen, one by one, to be discarded from $X$.
Then, the next element, $x_{t+1}$, is obtained as
\begin{equation}\label{eq:argmin_step_inductive}
    x_{t+1} = \operatorname*{arg\,min}_{x_r \in X \setminus \mathcal{R}_t}
    \left\| \Phi(x_r) - \mu_n + \varepsilon_t \right\|_2^2.
\end{equation}

The error vector and the removal set are then updated according to
\begin{equation}\label{eq:bkh_error_defintiniton}
    \varepsilon_{t+1} = \Phi(x_{t+1}) - \mu_n + \varepsilon_t,
    \quad
    \mathcal{R}_{t+1} = \mathcal{R}_t \cup \{x_{t+1}\}.
\end{equation}
%-------------------------------------------------------------------
\begin{algorithm}[t]
\caption{Backward Kernel Herding via Mean Embedding}\label{alg:backward_thinning}
\SetAlgoLined
\KwIn{Point set $X = \{x_1, \ldots, x_n\}$, feature map $\Phi:\R^d \longrightarrow \R^q$, 
      target size $T < n$}
\KwOut{Thinned set $X^* \subset X$ with $|X^*| = T$}

\BlankLine
Compute $\mu_n \leftarrow \dfrac{1}{n}\displaystyle\sum_{i=1}^{n} \Phi(x_i)$\;

\BlankLine
$\varepsilon \leftarrow 0 \in \mathcal{H}$\tcp*{Initialize residual error}
$X^{(0)} \leftarrow X$\;

\BlankLine
\For{$m = 0, 1, \ldots, n-T - 1$}{
    \BlankLine
    $r_m \leftarrow \displaystyle\operatorname*{arg\,min}_{r \in X^{(s)}}
    \left\| \Phi(x_r) - \mu_n - \varepsilon \right\|_2^2$\;

    \BlankLine
    $\varepsilon \leftarrow \Phi(x_{r_m}) - \mu_n + \varepsilon $\tcp*{Update residual error} 

    \BlankLine
    $X^{(m+1)} \leftarrow X^{(m)} \setminus \{x_{r_m}\}$\;
}

\BlankLine
\Return{$X^{(n-T)}$}
\end{algorithm}
The full procedure is summarized in Algorithm~\ref{alg:backward_thinning}.

If we do not want to rely in the explicit formulation of $\phi$ we can use the kernel trick: 
\begin{equation*}
    x_{t+1} = \operatorname*{arg\,min}_{x_r \in X \setminus \mathcal{R}_t}
    \left\| \Phi(x_r) - (\mu_n + \varepsilon_t) \right\|_2^2
    %\\
    = \operatorname*{arg\,min}_{x_r \in X \setminus \mathcal{R}_t}
    \left\| \Phi(x_r) \right\|_2^2 
    -2  \langle \Phi(x_r), \mu_n + \varepsilon_t \rangle
    +\left\|\mu_n + \varepsilon_t \right\|_2^2.
\end{equation*}
Note that the term $\left\|\mu_n + \varepsilon_t \right\|_2^2$ does not depend on $x_r$ so 
\begin{equation}
    x_{t+1} 
    = \operatorname*{arg\,min}_{x_r \in X \setminus \mathcal{R}_t}
    \left\| \Phi(x_r) \right\|_2^2 
    -2  \langle \Phi(x_r), \mu_n + \varepsilon_t \rangle
    +\left\|\mu_n + \varepsilon_t \right\|_2^2
    = 
    \operatorname*{arg\,min}_{x_r \in X \setminus \mathcal{R}_t}
    \left\| \Phi(x_r) \right\|_2^2 
    -2  \langle \Phi(x_r), \mu_n + \varepsilon_t \rangle.
\end{equation}
Using the kernel trick $\left\| \Phi(x_r) \right\|_2^2$ = $k(x_r, x_r)$
and 
%\begin{equation*}
$    \langle \Phi(x_r), \mu_n + \varepsilon_t \rangle 
    = 
     \langle \Phi(x_r), \mu_n  \rangle
     + 
      \langle \Phi(x_r),\varepsilon_t \rangle.
$%\end{equation*}
where 
\begin{equation*}
      \langle \Phi(x_r), \mu_n  \rangle = \frac{1}{n}\sum_{i = 1}
^n k(x_r, x_i)
\end{equation*}
The $\langle \Phi(x_r),\varepsilon_t \rangle$ could be defined inductively using \eqref{eq:bkh_error_defintiniton} and multiplying by $\phi(x_r)$ in each member of the equality
\begin{equation}
    \langle \varepsilon_{t+1} ,  \Phi(x_{r})\rangle
    =
    \langle \Phi(x_{r_{t+1}}),  \Phi(x_{r_{t}})\rangle
    - \langle \mu_n , \Phi(x_{r_{t+1}}) \rangle 
    + \langle \varepsilon_t , \Phi(x_{r_t})\rangle
    = 
    k(x_{r_{t+1}}, x_r) - \frac{1}{n}\sum_{i=1}^nk(x_r, x_i) +  \langle \varepsilon_{t} ,  \Phi(x_{r_{t}})\rangle.
\end{equation}
and for the base recursion
$\varepsilon_{1} = \phi(x_{r_1}) - \mu_n$ 
hence 
\begin{equation*}
    \langle \varepsilon_{1} ,  \Phi(x_{{r_1}})\rangle 
    = 
    k(x_{r_1}, x_{r_1}) - \frac{1}{n}\sum_{i=1}^nk(x_{r_1}, x_i).
\end{equation*}

\begin{algorithm}[t]
\caption{Kernelized Backward Herding via Mean Embedding}\label{alg:kernel_backward_herding_kernel_trick}
\SetAlgoLined
\KwIn{Point set $X = \{x_1, \ldots, x_n\}$, kernel function $k(\cdot, \cdot)$, 
      target size $T < n$}
\KwOut{Thinned set $X^* \subset X$ with $|X^*| = T$}

\BlankLine
\tcp{Precompute Kernel Mean Embeddings and initialize inner products}
\For{$x \in X$}{
    $K_\mu(x) \leftarrow \dfrac{1}{n}\displaystyle\sum_{i=1}^{n} k(x, x_i)$\;
    $I(x) \leftarrow K_\mu(x)$\; 
}

\BlankLine
$X^{(0)} \leftarrow X$\;

\BlankLine
\For{$m = 0, 1, \ldots, n-T - 1$}{
    \BlankLine
    \tcp{Find the point minimizing the kernelized objective}
    $x_{r_m} \leftarrow \displaystyle\operatorname*{arg\,min}_{x_r \in X^{(m)}}
    \left( k(x_r, x_r) - 2 I(x_r) \right)$\;

    \BlankLine
    $X^{(m+1)} \leftarrow X^{(m)} \setminus \{x_{r_m}\}$\;
    
    \BlankLine
    \tcp{Update inner products for remaining points}
    \For{$x \in X^{(m+1)}$}{
        $I(x) \leftarrow I(x) + k(x, x_{r_m}) - K_\mu(x)$\;
    }
}
\BlankLine
\Return{$X^{(n-T)}$}
\end{algorithm}
The kernel trick version of the method is described in Algorithm~\ref{alg:kernel_backward_herding_kernel_trick}.

\subsubsection{Observations}

\begin{description}
\item[Connection to Support Vector Machines.]
The selection criterion~\eqref{eq:argmin_step1} removes at each step the point 
$x_{r_m}$ whose feature map $\Phi(x_{r_m})$ lies closest to the mean embedding 
$\mu_n$ in $\mathcal{H}$. Geometrically, these are the points near the 
{center of mass} of the feature space, which correspond precisely to the 
interior points of the convex hull of $\{\Phi(x_i)\}_{i=1}^n$. This is 
complementary to SVM philosophy, where only the 
{support vectors}---the points farthest from the decision boundary, i.e., 
the most extreme points in feature space---are retained to define the solution. 
Backward Kernel Herding thus provides a principled justification for discarding 
non-support-like points first: the points removed are precisely those that an 
SVM would deem least informative, making the procedure a well-motivated 
heuristic for dataset compression.

\item[Complementarity with Kernel Herding.]
Let $\mu_n = \frac{1}{n} \sum_{i=1}^{n} \Phi(x_i)$ denote the target mean embedding of the full dataset $X$.
At iteration $m$, let $R_m = \{x_{r_0}, \dots, x_{r_{m-1}}\}$ be the set of $m$ discarded points, and let $X^{(m)} = X \setminus R_m$ be the remaining subset of size $n-m$.
The mean embedding of this remaining subset is defined as:
\begin{equation*}
    \mu^{(m)} = \frac{1}{n-m} \sum_{x \in X^{(m)}} \Phi(x)
\end{equation*}

From the definition of the Backward Kernel Herding, Algorithm~\ref{alg:backward_thinning},
the accumulated residual error $\varepsilon_m \in \mathcal{H}$ tracking the removed points satisfies:
\begin{equation*}
    \varepsilon_m = \sum_{x \in R_m} \Phi(x) - m\mu_n.
\end{equation*}
By conservation of the total dataset sum, the combined embeddings of the remaining and discarded subsets must equate to the full dataset embedding:
\begin{equation}\label{eq:n_mu}
    n\mu_n = \sum_{x \in X^{(m)}} \Phi(x) + \sum_{x \in R_m} \Phi(x).
\end{equation}
Substituting $\mu^{(m)}$ and $\varepsilon_m$ into~\eqref{eq:n_mu} yields:
\begin{equation*}
    n\mu_n = (n-m)\mu^{(m)} + (\varepsilon_m + m\mu_n).
\end{equation*}
Isolating the remaining mean embedding vector yields:
\begin{align*}
    n\mu_n - m\mu_n - \varepsilon_m &= (n-m)\mu^{(m)} \\
    (n-m)\mu_n - \varepsilon_m &= (n-m)\mu^{(m)} \\
    \mu_n - \mu^{(m)} &= \frac{\varepsilon_m}{n-m}.
\end{align*}
Taking norm $\|\cdot\|_{\mathcal{H}}^2$ on both sides yields the exact squared MMD error of the remaining dataset at step $m < n$:
\begin{equation*}
    \text{MMD}_{\text{BKH}}^2(X^{(m)}, \mu_n) = \left\| \mu^{(m)} - \mu_n \right\|_{\mathcal{H}}^2 = \frac{\|\varepsilon_m\|_{\mathcal{H}}^2}{(n-m)^2}.
\end{equation*}
For the MMD error formulation in Kernel Herding, we consider
Equation~(5) from~\cite{chen2012super},
\begin{equation*}
    \text{MMD}_{\text{KH}}^2(X^{(pn)}, \mu_n) = E_t = \left\|\mu_n - \frac{1}{T} \sum_{t=1}^T \phi(x_t)\right\|^2
    = 
    \left\| \mu^{(m)} - \mu_n \right\|_{\mathcal{H}}^2
    ,
\end{equation*}
where $T = \lfloor np \rfloor$
and corresponds to the MMD$^2$ error minimized. 
So the guarantees are the same for both methods.

An alternative argument is that both methods are greedy algorithms that optimize the same objective function, either through minimization or maximization, as noted by Bach~\cite{bach2012equivalence} in the context of Kernel Herding. The same interpretation applies to our Backward Kernel Herding construction.
\end{description}

\subsection{Methods comparison}
An important observation is that, although both Kernel Herding methods optimize the same objective, greedy algorithms are inherently path-dependent.
Consequently, even when the final coreset size is fixed, the resulting coresets may differ.

Table~\ref{tab:exact-complexity} compares the FLOPs and memory cost of all methods mentioned.
\begin{table}[t]
\centering
\caption{%
  Exact FLOPs and memory for each coreset construction method, derived
  directly from the algorithm complexity sections (not from summary tables).
  Notation: $n$ input size; $p \in (0,1)$ retention fraction so that the
  coreset size is $pn$ for all methods; $R$ number of KT-swap
  passes; $L = \lceil\log_2(1/\tau)\rceil$ binary expansion depth;
  $q$ RFF output dimension; $C_k$, $C_{k_\mathrm{rt}}$, $C_\Phi$ per-evaluation
  costs of the target kernel, square-root kernel, and feature map respectively.%
}
\label{tab:exact-complexity}
\renewcommand{\arraystretch}{2.4}
\begin{tabularx}{\textwidth}{%
  >{\raggedright\arraybackslash}X
  >{\centering\arraybackslash}X
  >{\centering\arraybackslash}X
  >{\centering\arraybackslash}X
  >{\centering\arraybackslash}X
}
\toprule
\textbf{Metric / Phase}
  & \textbf{Flexible Kernel Thinning (FKT)}
  & \textbf{Kernel Herding (KH)}
  & \textbf{Kernelized Backward Kernel Herding (Algorithm~\ref{alg:kernel_backward_herding_kernel_trick})}
  & \textbf{Backward Kernel Herding (Algorithm~\ref{alg:backward_thinning}, RFF)} \\
\midrule
Total FLOPs
  & $L\,n\,C_{k_\mathrm{rt}}$\newline
    $+\; p\,n^2\,C_{k_\mathrm{rt}}$\newline
    $+\; R\,p\,n^3\,C_k$
&  $n (1 + n C_k)$ \newline
$+ \frac{C_k}{6} (np+1)(24n + 3n^2p - 13np - 2n^2p^2)$ %\newline 
    & $n (1 + n C_k)$ \newline
    $+\frac{1}{2}(6+C_k)n(1-p)(np+1)$ 
  & $n\,C_\Phi + n\,q$\newline
    $+\;(C_\Phi+2q)\,\dfrac{(1-p)n\,((1-p)n+1)}{2}$ \\

\midrule

Total FLOPs ($\mathcal{O}$ notation)
  & $\mathcal{O}(R\,p\,n^3\,C_k)$
  & $\mathcal{O}(n^2\,C_k)$
  & $\mathcal{O}(n^2\,C_k)$
  & $\mathcal{O}(n^2\,C_\Phi)$ \\

\midrule

Memory
  & $\mathcal{O}(n)$\newline
    \small{(node index sets at each level;\newline no $n\times n$ kernel matrix)}
  & $2n + pn$ floats\newline
    \small{($\mathbf{b},\mathbf{r}^{(t)}\in\mathbb{R}^n$ + $pn$ coreset indices;\newline no $n\times n$ kernel matrix)}
  & $2n$ scalars\newline
    \small{($K_\mu(x)$ and $I(x)$\newline for all $x \in X$;\newline no $n\times n$ kernel matrix)}
  & $\mathcal{O}(nq)$\newline
    \small{($\{\Phi(x_i)\}_{i=1}^n \subset \mathbb{R}^q$\newline $+$ residual $\varepsilon_m \in \mathbb{R}^q$)} \\

\bottomrule
\end{tabularx}
\end{table}
To this end, let $p \in (0,1)$ denote the retention fraction, so that the coreset size is
$pn$ and the number of removal iterations is $(1-p)n$. From the exact FLOP
counts derived previously, the dominant terms are (using $\mathcal{F}$ to refer to these quantities)
\begin{align*}
  \mathcal{F}_{\mathrm{KH}}   &\approx \left(1 + p\right)n^2 C_k,\\
  \mathcal{F}_{\mathrm{KBKH}} &\approx \left(1 + \frac{(1-p)^2}{2}\right)n^2 C_k,\\
  \mathcal{F}_{\mathrm{BKH}}  &\approx \frac{(1-p)^2}{2}\,n^2 C_\Phi,\end{align*}
where $C_\Phi$ is the cost of a single RFF feature map evaluation.
FKT is excluded from the comparison as its KT-SWAP phase introduces a cubic
term $R\,p\,n^3 C_k$ that dominates in all regimes.

Kernelized BKH is cheaper than KH whenever
\begin{equation*}
  1 + \frac{(1-p)^2}{2} < 1 + p
  \;\iff\;
  p > 2 - \sqrt{3} \approx 0.27,
  \label{eq:crossover}
\end{equation*}
so the advantage of backward removal holds well below $p = 0.5$. For $p > 0.5$
the loop cost of Kernelized BKH is at most $\tfrac{1}{8}n^2 C_k$, whereas
that of KH exceeds $\tfrac{1}{2}n^2 C_k$, a reduction by more than a factor
of four. BKH with RFF further eliminates the fixed $\mathcal{O}(n^2 C_k)$
preprocessing of Kernelized BKH and replaces $C_k$ with $C_\Phi \leq C_k$,
making it strictly cheaper whenever $p > 2-\sqrt{3}$ and $C_k \geq C_\Phi$.

\medskip
As a consequence, we can state the following
\noindent\textbf{rule of thumb:}
KH is preferred when $p \leq 0.27$; Kernelized BKH should be used for $p \in (0.27, 1)$ when
an exact kernel is required; otherwise,  BKH with RFF is preferable, especially as
$p \to 1$ or when $C_k \gg C_\Phi$, since the loop cost vanishes as $(1-p)^2
\to 0$ and the $\mathcal{O}(n^2 C_k)$ preprocessing is entirely avoided.
% Finally, let us also provide a useful rule of thumb when deciding whether to use Kernel Herding or BKH.
% Since the approach is similar, one would prefer BKH when the size of the final subsample is over half of the dataset, since it will have a lower computational cost, and viceversa.

\section{Experiments}
\label{sec:experiments}

In this section we present the results of the experimental comparison of the performance of the algorithms we have discussed. We aim to compare the performance of Kernel Herding, Kernel Thinning and Backward Kernel Herding when selecting samples for supervised problems. 
The procedure followed in all the experiments is presented in Algorithm~\ref{alg:methodology}.
\begin{algorithm}
\SetAlgoLined
\DontPrintSemicolon
\SetKwInOut{Input}{Input}
\Input{Dataset $\mathcal{D}$, Selection Model $\mathcal{S}$, Supervised Kernel $\mathcal{K}$, Learning Model $\mathcal{M}$, Samples Percentage to retain, $p$}
\BlankLine
\For{$i \leftarrow 1$ \KwTo $10$}{
    \tcp{Step 1: Data Partitioning}
    Split $\mathcal{D}$ into $50\%$ training set $\mathcal{D}_{train}$ and $50\%$ test set $\mathcal{D}_{test}$ (using a common fixed seed)\;
    
    \BlankLine
    \tcp{Step 2: Coreset Selection}
    Apply $\mathcal{S}$ and $\mathcal{K}$ to $\mathcal{D}_{train}$ to extract $p\%$ of the samples for a representative coreset $\mathcal{C}$ (accounting for stochasticity in randomized methods)\;
    
    \BlankLine
    \tcp{Step 3: Model Training}
    Train the learning model $\mathcal{M}$ using the extracted coreset $\mathcal{C}$\;
    
    \BlankLine
    \tcp{Step 4: Evaluation}
    Compute prediction performance metrics and resource profiling on $\mathcal{D}_{test}$\;
}
\caption{Coreset Selection and Evaluation Framework \label{alg:methodology}}
\end{algorithm}

In order to do so, we conducted tests using two of the most prominent kernel-based learning paradigms: {Kernel Support Vector Machines (kSVM)} and {Gaussian Processes (GP)}~\cite{bishop2006pattern}. Both models are versatile frameworks capable of addressing classification and regression tasks by applying the \textit{kernel trick} to map input data into high-dimensional feature spaces.

Throughout all experiments, we employ the {Gaussian (RBF) kernel}.
To ensure reproducibility and eliminate manual tuning bias, the kernel hyperparameter $\gamma$ is determined via the heuristic commonly utilized in the \textit{scikit-learn}~\cite{scikit-learn} framework:
\begin{equation}
    \gamma = \frac{1}{n_{\text{features}} \cdot \text{Var}(X)}.
\end{equation}

In the specific case of regression via KSVM, the $\epsilon$-insensitivity parameter is computed using a data-driven heuristic to account for the noise scale of the target variable $y$:
\begin{equation}
    \epsilon = 0.1 \cdot \text{std}(y).
\end{equation}

% Modelos
% Datasets
We conducted the experiments with the OpenML benchmarking suites. This allows for varied datasets, which are widely adopted and have an emphasis on reproducibility. Specifically, we used the OpenML benchmarking suites introduced by \cite{bischl2017openml} for classification tasks and the curated OpenML-CTR23 suite proposed by~\cite{fischer2023openml} for regression tasks.
These benchmarks provide standardized datasets and evaluation settings, enabling fair and comparable assessment of model performance across a broad range of machine learning problems.

For evaluation, we used Balanced Accuracy as the primary metric for classification tasks, as it provides a robust assessment under class imbalance. For GP classification, we additionally report ROC-AUC because GP classifiers naturally yield probabilistic predictions. To avoid introducing additional calibration or score-transformation steps, ROC-AUC was not reported for KSVM models, and their performance was assessed using Balanced Accuracy only.

For all regression tasks, performance was measured using the coefficient of determination, $R^2$.
We also measure the training time and memory consumption. Training time is defined as the total time required for coreset construction and model training (Steps 2 and 3 in Algorithm~\ref{alg:methodology}).
Similarly, memory consumption corresponds to the total memory footprint, measured in MB, required during these two stages.
These metrics complement the theoretical analysis presented in Table~\ref{tab:exact-complexity}.
While empirical measurements capture the practical behavior of the implementations, Table~\ref{tab:exact-complexity} provides the corresponding FLOPs and asymptotic computational complexity estimates only for the subsampling methods without taking into account the learning methods.

For each dataset, we evaluate model performance using the metrics generated from the $10$ outer folds of the cross-validation procedure presented in Algorithm~\ref{alg:methodology}.
To rigorously compare each pair of models $(\mathcal{V}_i, \mathcal{V}_j)$, we employ the \textit{Wilcoxon signed-rank test} \cite{wilcoxon1945individual} on the fold-wise performance differences. Since comparing $k$ models involves $k(k-1)/2$ pairwise tests, we adjust the resulting $p$-values using the \textit{Holm--Bonferroni correction}~\cite{Holm1979} to control the family-wise error rate at a significance level of $\alpha = 0.05$.

A model $\mathcal{V}_i$ is considered significantly better than model $\mathcal{V}_j$ if it exhibits a higher mean performance and the corresponding adjusted $p$-value is strictly less than $\alpha$.
Based on these outcomes, we construct a dataset-specific ranking using a \textit{competition ranking scheme}: the rank assigned to a model is $1 + n_{better}$, where $n_{better}$ represents the number of models that are statistically superior to it.
In cases where no significant difference is detected, models share the same rank, inherently accounting for statistical ties.

To synthesize findings across the entire benchmark, we calculate the \textit{Average Rank} for every model across all $N$ datasets. This metric serves as a robust indicator of global relative performance, minimizing the influence of disparities between datasets.

To determine if the observed differences in average ranks are statistically significant, we follow the framework established by~\cite{Demsar2006}.
We first apply the \textit{Friedman test} to evaluate the global null hypothesis. If rejected, we conduct the \textit{Nemenyi post-hoc test}, defining the \textit{Critical Distance} (CD) as:
$$CD = q_{\alpha, k} \sqrt{\frac{k(k+1)}{6N}}$$
where $q_{\alpha, k}$ is the critical value based on the studentized range statistic, $k = 3$ the number of models and $N = 72$ in classification and $N = 35$ for classification.
Two models are identified as significantly different if the absolute difference between their average ranks exceeds the CD. This two-step procedure ensures that our claims of model superiority are statistically grounded, accounting for both dataset-level noise and the multiplicity of the experimental design. 
In our case, due to the benchmarks sizes, for $\alpha= 0.05$ and classification benchmark the CD is  $0.391$ and for the regression benchmark $0.560$.
We did not place all the results for every dataset in this document due to space constrains, but all the experiment sources are available via~\url{https://github.com/BlancaCC/backward-thinning}.

\begin{table}[t]
\centering
\caption{Predictive error ranking (the lower the better) across benchmark datasets. Best results and those statistically indistinguishable from the best ($p > 0.05$ via Nemenyi CD) are highlighted in bold.}
\label{tab:predictive_error}
\renewcommand{\arraystretch}{1.15}
\begin{tabular}{llccc}
\hline
\textbf{Task} & \textbf{Method} & \textbf{Size 25\%} & \textbf{Size 50\%} & \textbf{Size 75\%} \\
\hline
\multirow{3}{*}{KSVM Classification}
& KH & 1.60 $\pm$ 0.69 & 1.75 $\pm$ 0.63 & 1.61 $\pm$ 0.55 \\
& FKT & \textbf{1.03 $\pm$ 0.24} & \textbf{1.09 $\pm$ 0.33} & \textbf{1.07 $\pm$ 0.31} \\
& BKH & 1.69 $\pm$ 0.75 & 1.81 $\pm$ 0.67 & 1.69 $\pm$ 0.65 \\ \hline
\multirow{3}{*}{GP Classification}
& KH & \textbf{1.58 $\pm$ 0.68} & 1.88 $\pm$ 0.73 & 1.67 $\pm$ 0.73 \\
& FKT & \textbf{1.41 $\pm$ 0.74} & \textbf{1.19 $\pm$ 0.51} & \textbf{1.10 $\pm$ 0.40} \\
& BKH & \textbf{1.38 $\pm$ 0.58} & 1.69 $\pm$ 0.65 & 1.60 $\pm$ 0.62 \\ \hline
\multirow{3}{*}{KSVM Regression}
& KH & \textbf{1.68 $\pm$ 0.79} & 2.10 $\pm$ 0.60 & 2.16 $\pm$ 0.64 \\
& FKT & \textbf{1.58 $\pm$ 0.81} & \textbf{1.10 $\pm$ 0.30} & \textbf{1.03 $\pm$ 0.18} \\
& BKH & \textbf{1.55 $\pm$ 0.68} & 1.97 $\pm$ 0.71 & 2.06 $\pm$ 0.63 \\ \hline
\multirow{3}{*}{GP Regression}
& KH & \textbf{1.32 $\pm$ 0.60} & \textbf{1.41 $\pm$ 0.63} & \textbf{1.22 $\pm$ 0.42} \\
& FKT & \textbf{1.26 $\pm$ 0.58} & \textbf{1.24 $\pm$ 0.58} & \textbf{1.35 $\pm$ 0.65} \\
& BKH & \textbf{1.32 $\pm$ 0.65} & \textbf{1.52 $\pm$ 0.74} & \textbf{1.48 $\pm$ 0.73} \\
\hline
\end{tabular}
\end{table}

\begin{table}[t]
\centering
\caption{Training time ranking (the lower the better) across benchmark datasets. Best results and those statistically indistinguishable from the best ($p > 0.05$ via Nemenyi CD) are highlighted in bold.}
\label{tab:training_time}
\renewcommand{\arraystretch}{1.15}
\begin{tabular}{llccc}
\hline
\textbf{Task} & \textbf{Method} & \textbf{Size 25\%} & \textbf{Size 50\%} & \textbf{Size 75\%} \\
\hline
\multirow{3}{*}{KSVM Classification}
& KH & \textbf{1.00 $\pm$ 0.00} & \textbf{1.00 $\pm$ 0.00} & 1.63 $\pm$ 0.49 \\
& FKT & 3.00 $\pm$ 0.00 & 3.00 $\pm$ 0.00 & 3.00 $\pm$ 0.00 \\
& BKH & 2.00 $\pm$ 0.00 & 1.84 $\pm$ 0.37 & \textbf{1.19 $\pm$ 0.39} \\ \hline
\multirow{3}{*}{GP Classification}
& KH & \textbf{1.09 $\pm$ 0.29} & \textbf{1.09 $\pm$ 0.28} & \textbf{1.10 $\pm$ 0.35} \\
& FKT & 2.95 $\pm$ 0.21 & 2.52 $\pm$ 0.78 & 2.30 $\pm$ 0.93 \\
& BKH & \textbf{1.42 $\pm$ 0.50} & \textbf{1.16 $\pm$ 0.37} & \textbf{1.17 $\pm$ 0.38} \\ \hline
\multirow{3}{*}{KSVM Regression}
& KH & \textbf{1.00 $\pm$ 0.00} & \textbf{1.00 $\pm$ 0.00} & \textbf{1.42 $\pm$ 0.50} \\
& FKT & 3.00 $\pm$ 0.00 & 3.00 $\pm$ 0.00 & 3.00 $\pm$ 0.00 \\
& BKH & 2.00 $\pm$ 0.00 & 1.94 $\pm$ 0.25 & \textbf{1.42 $\pm$ 0.50} \\ \hline
\multirow{3}{*}{GP Regression}
& KH & \textbf{1.06 $\pm$ 0.25} & \textbf{1.17 $\pm$ 0.38} & \textbf{1.22 $\pm$ 0.42} \\
& FKT & 3.00 $\pm$ 0.00 & 2.31 $\pm$ 0.89 & 2.17 $\pm$ 0.94 \\
& BKH & \textbf{1.04 $\pm$ 0.21} & \textbf{1.14 $\pm$ 0.44} & \textbf{1.29 $\pm$ 0.46} \\
\hline
\end{tabular}
\end{table}

\begin{table}[t]
\centering
\caption{Memory usage (the lower the best) across benchmark datasets. Best results and those statistically indistinguishable from the best ($p > 0.05$ via Nemenyi CD) are highlighted in bold.}
\label{tab:memory_usage}
\renewcommand{\arraystretch}{1.15}
\begin{tabular}{llccc}
\hline
\textbf{Task} & \textbf{Method} & \textbf{Size 25\%} & \textbf{Size 50\%} & \textbf{Size 75\%} \\
\hline
\multirow{3}{*}{KSVM Classification}
& KH & 2.59 $\pm$ 0.50 & 2.57 $\pm$ 0.50 & 2.53 $\pm$ 0.50 \\
& FKT & \textbf{1.00 $\pm$ 0.00} & \textbf{1.00 $\pm$ 0.00} & \textbf{1.00 $\pm$ 0.00} \\
& BKH & 2.00 $\pm$ 0.00 & 2.00 $\pm$ 0.00 & 2.03 $\pm$ 0.17 \\ \hline
\multirow{3}{*}{GP Classification}
& KH & 2.45 $\pm$ 0.56 & 2.07 $\pm$ 0.53 & 2.28 $\pm$ 0.52 \\
& FKT & \textbf{1.03 $\pm$ 0.17} & 1.88 $\pm$ 0.33 & 1.73 $\pm$ 0.45 \\
& BKH & 1.95 $\pm$ 0.21 & \textbf{1.03 $\pm$ 0.18} & \textbf{1.23 $\pm$ 0.43} \\ \hline
\multirow{3}{*}{KSVM Regression}
& KH & 3.00 $\pm$ 0.00 & 3.00 $\pm$ 0.00 & 3.00 $\pm$ 0.00 \\
& FKT & \textbf{1.00 $\pm$ 0.00} & \textbf{1.00 $\pm$ 0.00} & \textbf{1.00 $\pm$ 0.00} \\
& BKH & 2.00 $\pm$ 0.00 & 2.00 $\pm$ 0.00 & 2.00 $\pm$ 0.00 \\ \hline
\multirow{3}{*}{GP Regression}
& KH & 2.97 $\pm$ 0.18 & 2.31 $\pm$ 0.66 & 2.17 $\pm$ 0.58 \\
& FKT & \textbf{1.00 $\pm$ 0.00} & \textbf{1.66 $\pm$ 0.55} & 1.74 $\pm$ 0.45 \\
& BKH & 1.97 $\pm$ 0.18 & \textbf{1.28 $\pm$ 0.45} & \textbf{1.17 $\pm$ 0.39} \\
\hline
\end{tabular}
\end{table}

Table~\ref{tab:predictive_error} reports the average ranking of the predictive error obtained by the different dataset reduction strategies across all benchmark datasets. 
Lower values indicate better predictive performance.
The reported ranks summarize the relative performance of each method, while the Nemenyi post-hoc test identifies methods that are not significantly different from the best-performing approach.

Overall, FKT consistently achieves the best predictive performance across most tasks and reduction sizes.
In KSVM classification, it clearly outperforms both KH and BKH for all reduction levels, obtaining the lowest average rank and being the only method consistently highlighted as statistically equivalent to the best result.
This finding suggests that theoretical guarantees have an impact in the prediction performance.

A similar trend can be observed for GP classification and KSVM regression.
For moderate reductions (50\% and 75\% of the original data retained), FKT substantially improves predictive performance with respect to KH and BKH.
Interestingly, when the reduction is more aggressive (25\% retained), the differences become less pronounced and all methods are statistically indistinguishable.
This behavior is consistent with the theoretical properties of kernel thinning. While the supervised kernel provides stronger guidance during the selection process, aggressive reductions inevitably degrade the quality of the approximation of the original empirical distribution, reducing the advantage of the supervised criterion.
Nevertheless, even under these extreme compression ratios, FKT remains among the best-performing methods.

The only scenario where the superiority of FKT is less evident corresponds to GP regression. In this task, all methods obtain statistically equivalent results for every reduction size.
In fact, no clear winner emerges from the comparison.
One possible explanation is that the supervised kernels employed in this study may not fully capture the characteristics required by Gaussian Processes.
However, the strong improvements observed for KSVM regression indicate that the supervised kernel construction itself is capable of enhancing regression performance.
Therefore, a more plausible explanation is that the probabilistic nature of Gaussian Processes interacts differently with the selected subsets, suggesting that specialized thinning or coreset construction procedures tailored to GPs may be required. Investigating this interaction constitutes an interesting direction for future work.

The computational efficiency results are summarized in Table~\ref{tab:training_time}, which reports the average ranking of the training time, which was measured via a hardware profiler built on PyAPI.
Lower ranks indicate faster model training.
As anticipated from the theoretical analysis, BKH tends to provide the most favorable trade-off between compression quality and subset construction cost.
In particular, for KSVM classification with a 75\% retained subset, BKH achieves the best ranking and significantly outperforms FKT.
This result is expected since BKH directly removes points from the complete dataset and avoids the iterative optimization required by KT.
For the remaining tasks, BKH and KH are generally statistically indistinguishable, both substantially outperforming FKT.
The latter consistently obtains the highest training-time ranks because the supervised kernel evaluation and iterative thinning procedure introduce an additional computational overhead during the subset construction stage.
Nevertheless, it is important to emphasize that this additional cost is compensated by the substantial predictive improvements observed in Table~\ref{tab:predictive_error}, particularly for classification tasks and support vector regression.

Finally, Table~\ref{tab:memory_usage} presents the average ranking of memory consumption, which was measured via a hardware profiler built on PyAPI. The results reveal a markedly different behavior from that observed for training time.
In KSVM classification and regression, FKT consistently achieves the lowest memory usage across all reduction sizes.
This outcome is particularly relevant because it demonstrates that not only improves predictive performance but also yields compact representations that require less memory during model training.

For GP classification, the comparison is more balanced. FKT provides the best memory efficiency for the strongest reduction level, whereas BKH becomes competitive as the retained subset size increases. Both methods outperform KH, indicating that the reduction strategy itself plays a more important role than the specific selection criterion in controlling memory requirements for GP classifiers.

A particularly interesting observation arises in GP regression. Although BKH achieves the best average ranks for the larger retained subsets, the differences with respect to FKT are not statistically significant. This behavior can be explained by the implementation characteristics of Gaussian Process models. Besides storing the selected subset, GP regression also requires maintaining additional model-specific structures whose memory footprint can dominate the total consumption.
Consequently, the relative differences introduced by the reduction strategy become less pronounced, making it difficult to distinguish statistically between FKT and BKH.

Taken together, the three tables indicate that FKT provides the best predictive performance in most scenarios, especially for classification tasks and KSVM regression, while maintaining competitive memory requirements.
BKH and KH remain attractive alternatives when computational efficiency is the primary concern, particularly regarding training time, and BKH is the best option if the reduction is smaller than $50\%$.
Therefore, the choice between methods depends on the desired trade-off between predictive accuracy and computational resources.
However, when predictive performance is the main objective, the results strongly support the use of FKT as the preferred reduction strategy.

\section{Conclusions and further work}
\label{sec:conclusions}
In this paper, we revisited kernel subsampling methods, presented a formal supervised formulation for a procedure that is usually omitted altogether, and complemented this study with a software contribution. Additionally, we proposed two new algorithms in this line.

First, we introduced Backward Kernel Herding (BKH), a novel subset selection framework designed to mitigate the high computational cost that limits the scalability of kernel-based methods.
Unlike traditional forward-construction techniques such as Kernel Herding, our approach operates via backward propagation, starting with the complete dataset and iteratively removing points to minimize the distortion of the mean embedding.
A primary advantage of this framework is that it reduces the number of greedy iterations required in realistic scenarios.
In particular, it accelerates the subsampling process when the desired reduction is smaller than $27\%$ of the dataset.
At the same time, it achieves predictive performance comparable to that of the state-of-the-art methods.
Furthermore, we explored BKT's links to existing literature, through its relation to KH.
And we also provided a principled justification of the suitability of the method for Support Vector Machines, since the method works by systematically discarding the least informative, ``interior'' points first, that may not be support vectors.

Secondly, we proposed Flexible Kernel Thinning (FKT), a variant of Kernel Thinning (KT) that overcomes the rigid power-of-two size constraints of the standard formulation. By exploiting the binary expansion of the target retention fraction, FKT allows for the construction of coresets of arbitrary size while preserving the computational cost and theoretical guarantees of KT. Additionally, we extended both existing and novel subset selection techniques to supervised learning settings. Incorporating joint and supervised kernels ensures that label dependencies are not ignored during the compression of classification and regression datasets.

Finally, our comprehensive empirical evaluation confirms the practical viability of the proposed methods.
The results show that BKH consistently matches the performance of state-of-the-art reduction techniques, including KH and KT, while often providing favorable computational properties, particularly in terms of training time.
More importantly, the proposed FKT variant frequently achieves the best predictive performance across classification and regression tasks, especially under moderate compression ratios, while maintaining competitive memory requirements.
Overall, this work broadens the landscape of kernel-based coreset construction methods by introducing both a competitive backward selection strategy and a size-flexible thinning framework, offering practitioners a wider range of accuracy–efficiency trade-offs for large-scale learning problems. Furthermore, it also delves into a supervised framework.

A possible direction for future work is the construction of ensemble methods based on different coresets.
For instance, once the kernel means and transformation values have been computed, they could be reused for both BKH and KH in order to generate two different coresets.
Since these coresets are generally not identical, combining them through an ensemble strategy could improve performance and robustness.

Moreover, all the methods considered in this work are based on greedy algorithms that select or remove one element at a time.
It could, therefore, be interesting to investigate strategies that process multiple elements simultaneously, for example selecting or removing two points at each iteration. Such an approach might improve the final approximation quality, although it would also introduce additional computational cost.

% --- Acknowledgments ---
\section*{Acknowledgments}
The authors acknowledge financial support from project PID2022-139856NB-I00 funded by MCIN/ AEI / 10.13039/501100011033 / FEDER, UE, project Inteligencia artificial para la industria 4.0: generación de datos, modelado avanzado, optimización e interpretabilidad (TEC-2024/COM-89) from the Autonomous Community of Madrid, and project IDEA-CM (TEC-2024/COM-89) from the Autonomous Community of Madrid and from the ELLIS Unit Madrid, Cátedra UAM-IIC de Ciencia de Datos y Aprendizaje Automático and FPI-UAM.

The authors acknowledge computational support from the Centro de Computación Científica-Universidad Autónoma de Madrid (CCC-UAM). The authors thank Estrella Sánchez for valuable discussions and support.

% --- CRediT Statement ---
\section*{CRediT authorship contribution statement}
\textbf{Blanca Cano-Camarero:} Conceptualization, Methodology, Software, Validation, Formal analysis, Investigation, Data curation, Writing - original draft, Visualization, Project administration. \\
\textbf{Yago R. Aguado-Carrillo-de-Albornoz:} Conceptualization, Methodology, Software, Validation, Formal analysis, Investigation, Data curation, Writing - original draft. \\
\textbf{Ángela Fernández-Pascual:} Conceptualization, Supervision, Writing - review and editing. \\
\textbf{José R. Dorronsoro:} Conceptualization, Supervision, Writing - review and editing.

% --- AI Declaration ---
\section*{Declaration of Generative AI and AI-assisted technologies in the writing process}
During the preparation of this work the authors used ChatGPT in order to improve the language style. After using this tool/service, the authors reviewed and edited the content as needed and take full responsibility for the content of the publication.

% --- Bibliography ---
\printbibliography

\end{document}